\documentclass{article}
\usepackage{spconf}
\usepackage{epsfig}
\usepackage{graphicx}
\usepackage{amsmath,amsfonts,amsthm,amssymb}
\usepackage{subcaption,cite}
\usepackage{algorithm,balance}
\usepackage{algpseudocode}
\usepackage{algorithmicx}
\usepackage{float}
\usepackage{adjustbox}
\usepackage{multirow,capt-of, etoolbox}

\newcommand{\etal}{et al.~\!}
\newcommand{\eg}{e.g~\!}
\newcommand\textlcsc[1]{\textsc{\MakeLowercase{#1}}}

\makeatletter
\newenvironment{chapquote}[2][2em]
  {\setlength{\@tempdima}{#1}%
   \def\chapquote@author{#2}%
   \parshape 1 \@tempdima \dimexpr\textwidth-2\@tempdima\relax%
   \itshape}
  {\par\normalfont\hfill--\ \chapquote@author\hspace*{\@tempdima}\par\bigskip}
\makeatother

\title{A\textlcsc{rt}GAN: Artwork Synthesis with Conditional Categorical GAN\textlcsc{s}}

\name{Wei Ren Tan$^{\star}$ \qquad Chee Seng Chan$^{\ddagger}$ \qquad Hern\'an E. Aguirre$^{\star}$ \qquad Kiyoshi Tanaka$^{\star}$}
\address{$^{\star}$Faculty of Engineering, Shinshu University, Nagano, Japan \\ $^{\ddagger}$Centre of Image \& Signal Processing, Fac. Comp. Sci. \& Info. Tech., University of Malaya, Malaysia\\
\{{\it 14st203c@shinshu-u.ac.jp; cs.chan@um.edu.my; ahernan@shinshu-u.ac.jp; ktanaka@shinshu-u.ac.jp\}}}

\begin{document}

\makeatletter
\let\@oldmaketitle\@maketitle
\renewcommand{\@maketitle}{\@oldmaketitle
\centering
\captionof{figure}{Artwork Generation: Comparison between DCGAN (top), GAN/VAE (middle), and A\textlcsc{rt}GAN (bottom)}
\label{fig:clscom}
  \includegraphics[width=0.31\linewidth,height=9.5\baselineskip]
    {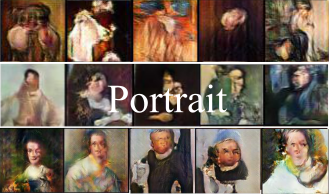}\quad\includegraphics[width=0.31\linewidth,height=9.5\baselineskip]
    {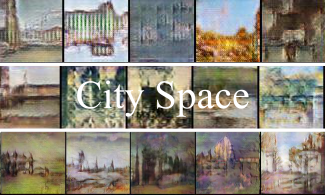}\quad\includegraphics[width=0.31\linewidth,height=9.5\baselineskip]
    {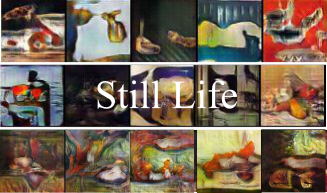}\bigskip}
\makeatother

\maketitle

\begin{abstract}

This paper proposes an extension to the Generative Adversarial Networks (GANs), namely as A\textlcsc{rt}GAN to synthetically generate more challenging and complex images such as artwork that have abstract characteristics. This is in contrast to most of the current solutions that focused on generating natural images such as room interiors, birds, flowers and faces. The key innovation of our work is to allow back-propagation of the loss function w.r.t. the labels (randomly assigned to each generated images) to the generator from the discriminator. With the feedback from the label information, the generator is able to learn faster and achieve better generated image quality. Empirically, we show that the proposed A\textlcsc{rt}GAN is capable to create realistic artwork, as well as generate compelling real world images that globally look natural with clear shape on CIFAR-10.
\end{abstract}
\begin{keywords}
image synthesis, generative adversarial networks, deep learning
\end{keywords}

\section{Introduction}

\begin{chapquote}{Pablo Picasso}
``I paint objects as I think them, not as I see them''
\end{chapquote}

Recently, Generative Adversarial Networks (GANs) \cite{goodfellow2014generative,salimans2016improved,denton2015deep,mirza2014conditional} have shown significant promise in synthetically generate natural images using the MNIST \cite{lecun1998mnist}, CIFAR-10 \cite{krizhevsky2009learning}, CUB-200 \cite{WelinderEtal2010} and LFW datasets \cite{LFWTechUpdate}. However, we could notice that all these datasets have some common characteristics: i) Most of the background/foreground are clearly distinguishable;  ii) Most of the images contain only one object per image and finally iii) Most of the objects have fairly structured shape such as numeric, vehicles, birds, face etc.

\begin{figure*}[ht]
\centering
\includegraphics[height=0.36\linewidth, width=0.95\linewidth]{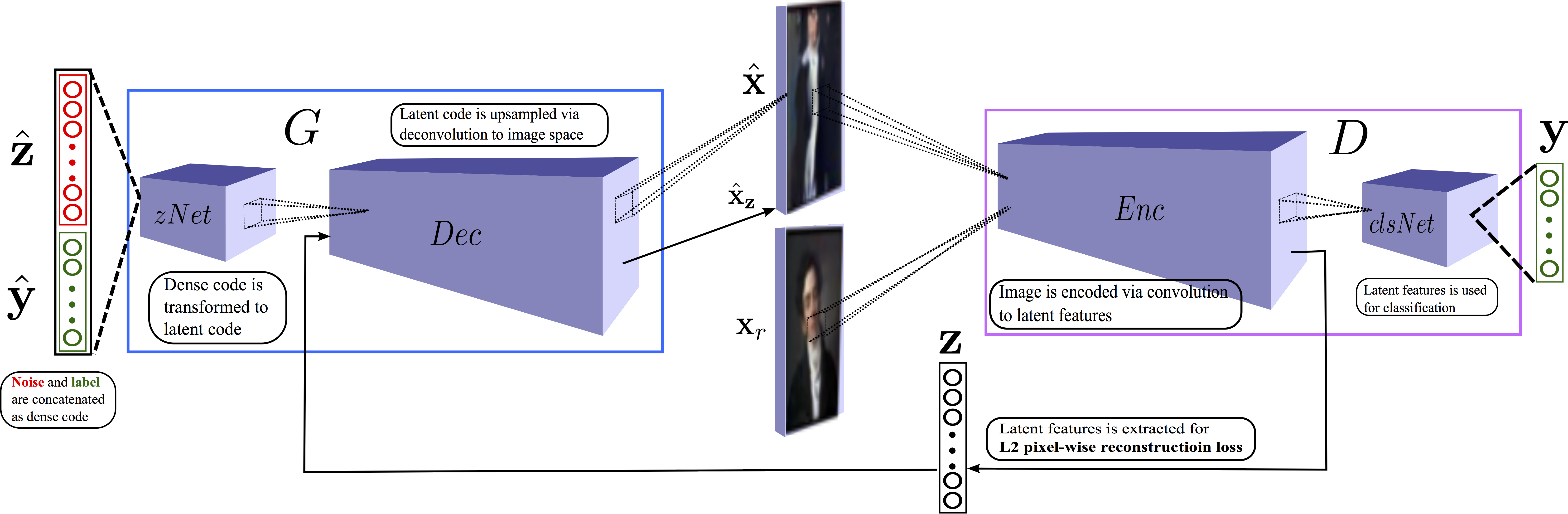}
\caption{The overall architecture of A\textlcsc{rt}GAN. The overall design is similar to the standard GAN except that we have additional input labels $\hat{\mathbf{y}}$ for $G$ and $D$ outputs probability distribution of labels. A connection is also added from $Enc$ to $Dec$ to reconstruct image for L2 pixel-wise reconstruction loss.}
\label{fig:ccgan}
\end{figure*}

In this paper, we would like to investigate if machine can create (more challenging) images that do not exhibit any of the above characteristics, such as the artwork depicted in Fig. \ref{fig:clscom}. Artwork is a mode of creative expression, coming in different kind of forms, including drawing, naturalistic, abstraction, etc. For instance, artwork can be non-figurative nor representable, \eg~{\it Abstract} paintings. Therefore, it is very hard to understand the background/foreground in the artwork. In addition, some artwork do not follow natural shapes, \eg~{\it Cubism} paintings. In the philosophy of art, aesthetic judgement is always applied to artwork based on one's sentiment and taste, which shows one's appreciation of beauty.

An artist teacher wrote an online article \cite{arturl} and pointed out that an effective learning in art domain requires one to focus on a particular type of skills (\eg~practice to draw a particular object or one kind of movement) at a time. Meanwhile, the learning in GANs only involves unlabeled data that doesn't necessarily reflect on a particular subject. In order to imitate such learning pattern, we propose to train GANs focuses on a particular subject by inputting some additional information to it. A similar approach is the Conditional GANs (CondGAN) \cite{mirza2014conditional}. The work feed a vector $\vec{y}$ into $D$ and $G$ as an additional input layer. However, there is no feedback from $\vec{y}$ to the intermediate layers. A natural extension is to train $D$ as a classifier with respect to $\vec{y}$ alike to the Categorical GANs (CatGAN) \cite{springenberg2015unsupervised} and Salimans \etal \cite{salimans2016improved}. In the former, the work extended $D$ in GANs to $K$ classes, instead of a binary output. Then, they trained the CatGAN by either minimize or maximize the Shannon entropy to control the uncertainty of $D$. In the latter, the work proposed a semi-supervised learning framework and used $K+1$ classes with an additional FAKE class. An advantage of such design is that it can be extended to include more (adversarial) classes, \eg Introspective Adversarial Networks (IAN) \cite{brock2016neural} used a ternary adversarial loss that forces $D$ to label a sample as reconstructed in addition to real or fake. However, such work do not use the information from the labels to train $G$. 

To this end, we propose a novel adversarial networks namely as A\textlcsc{rt}GAN that is close to CondGAN \cite{mirza2014conditional} but it differs in such a way that we feed $\vec{y}$ to $G$ only and back-propagate errors to $G$. This allows $G$ to learn better by using the feedback information from the labels. At the same time, A\textlcsc{rt}GAN outputs $K+1$ classes in $D$ as to the \cite{springenberg2015unsupervised} but again we differ in two ways: First, we set a label to each generated images in $D$ based on $\vec{y}$. Secondly, we use sigmoid function instead of softmax function in $D$. This generalizes the A\textlcsc{rt}GAN architecture so that it can be extended to other works, \eg~multi-labels problem \cite{boutell2004learning}, open set recognition problem \cite{scheirer2014TPAMIb}, etc. Inspired by Larsen \etal \cite{larsen2015autoencoding}, we also added the L2 pixel-wise reconstruction loss along with the adversarial loss to train $G$ in order to improve the quality of the generated images. Empirically, we show qualitatively that our model is capable to synthesize descent quality artwork that exhibit for instance famous artist styles such as Vincent van Vogh (Fig. \ref{vangogh2}). At the same time, our model also able to create samples on CIFAR-10 that look more natural and contain clear object structures in them, compared to DCGAN \cite{radford2015unsupervised} (Fig. \ref{fig:cifar}). 

\section{Approach}
\label{secP}

In this section, we present a novel framework built on GANs  \cite{goodfellow2014generative}. We begin with a brief concept of the GANs framework. Then, we introduce the A\textlcsc{rt}GAN.

\subsection{Preliminaries}

The GANs framework \cite{goodfellow2014generative} was established with two competitors, the Generator $G$ and Discriminator $D$. The task of $D$ is to distinguish the samples from $G$ and training data. While, $G$ is to confuse $D$ by generating samples with distribution close to the training data distribution. The GANs objective function is given by:
{\small
\begin{equation}
\min_G\max_D (\mathbb{E}_{\mathbf{x}\sim p_{data}}\log p(\mathbf{y}|\mathbf{x})+\mathbb{E}_{\hat{\mathbf{z}}\sim p_{noise}}[1-\log p(\mathbf{y}|G(\hat{\mathbf{z}}))])
\label{eq:gan}
\end{equation}}
\noindent where $D$ is trained by maximizing the probability of the training data (first term), while minimizing the probability of the samples from $G$ (second term). 

\subsection{A\textlcsc{rt}GAN}

The basic structure of A\textlcsc{rt}GAN is similar to GANs: it consists of a discriminator and a generator that are simultaneously trained using the minmax formulation of GANs, as described in Eq. \ref{eq:gan}. The key innovation of our work is to allow feedback from the labels given to each generated image through the loss function in $D$ to $G$. That is, we feed additional (label) information $\hat{\mathbf{y}}$ to the GANs network to imitate how human learn to draw. This is almost similar to the CondGAN \cite{mirza2014conditional} which is an extension of the GANs in which both $D$ and $G$ receive an additional vector of information $\hat{\mathbf{y}}$ as input. That is, $\hat{\mathbf{y}}$ encodes the information of either the attributes or classes of the data to control the modes of the data to be generated. However, it has one limitation as the information of $\hat{\mathbf{y}}$ is not fully utilized through the back-propagation process to improve the quality of the generated images. Therefore, a natural refinement is to train $D$ as a classifier with respect to $\hat{\mathbf{y}}$. To this end, we modify $D$ to output probability distribution of the labels, as to CatGAN \cite{springenberg2015unsupervised} except that we set a label to each generated images in $D$ based on $\hat{\mathbf{y}}$ and use cross entropy to back-propagate the error to $G$. This allows $G$ to learn better by using the feedback information from the labels.  Conceptually, this step not only help in speeding up the training process, but also assists the A\textlcsc{rt}GAN to grasp more abstract concepts, such as artistic styles which are crucial when generating fine art paintings. Also, we use sigmoid function instead of softmax function in $D$, and employ an additional L2 pixel-wise reconstruction loss as to Larsen \etal~\cite{larsen2015autoencoding} along with adversarial loss to improve the training stability. Contrast to Larsen \etal \cite{larsen2015autoencoding}, in A\textlcsc{rt}GAN architecture, the Decoder $D$ shares the same network with Encoder $Enc$ only.

\subsection{Details and Formulation of Architecture}

Fig. \ref{fig:ccgan} depicts the overall architecture of the proposed A\textlcsc{rt}GAN. Formally, $D$ maps an input image $\mathbf{x}$ to a probability distribution $p(\mathbf{y}|\mathbf{x})$, $D:\mathbf{x}\rightarrow p(\mathbf{y}|\mathbf{x})$. Generally, $D$ can be separated into two parts: an encoder $Enc$ that produces a latent feature $\mathbf{z}$ followed by a classifier $clsNet$. Similarly, $G$ is fed with a random vector $\hat{\mathbf{z}} \in {\mathbb{R}^d} \sim \mathcal{N}(0,1)^d$ concatenated with the label information $\hat{\vec{y}}$ and outputs a generated image $\hat{\mathbf{x}}$, such that $G: [\hat{\mathbf{z}}, \hat{\vec{y}}]\rightarrow \hat{\mathbf{x}}$. $G$ composes of a $zNet$ that transforms the input to a latent space, followed by a decoder $Dec$. In this context, $\mathcal{N}(0, 1)$ is a normal-distributed random number generator with mean $0$ and standard standard deviation of $1$, and $d$ is the number of elements in $\hat{\mathbf{z}}$.

Given $K$ labels and $K+1$ representing the FAKE class, $\mathbf{y}_k$ and $\hat{\mathbf{y}}_{\hat{k}}$ are denoted as one-hot vectors, such that $\mathbf{y}=[y_1,y_2,\ldots,y_{K+1}]$ and $\hat{\mathbf{y}}=[\hat{y}_1,\hat{y}_2,\ldots,\hat{y}_{K}]$, where $y_{k},\hat{y}_{\hat{k}}=1$ and $y_{i\setminus k},\hat{y}_{j\setminus\hat{k}}=0$, $i=\{1,2,\ldots,K+1\}$, $j=\{1,2,\ldots,K\}$ when $k$ and $\hat{k}$ are the true classes of the real and generated images, respectively. Then, the data draw from the real distribution is denoted $(\mathbf{x}_r, k)\sim p_{data}$, where $k\in\mathbf{K}=\{1,2,\ldots,K\}$ is the label of $\mathbf{x}_r$. Meanwhile, $p_{noise}$ is the noise distribution for $\hat{\mathbf{z}}$. For simplicity, we use $G(\hat{\mathbf{z}}, \hat{\mathbf{y}}_{\hat{k}})$ to express the output of $G$, such that $\hat{k}$ is randomly chosen. Hence, we can minimize the loss function, $\mathcal{L}_{D}$ w.r.t parameters $\theta_D$ in $D$ to update $D$:
{\small \begin{align}
\mathcal{L}_{D} = &- \mathbb{E}_{(\mathbf{x}_r, k)\sim p_{data}}\big[\log p(y_i|\mathbf{x}_r, i=k) \nonumber \\
&+ \log (1-p(y_i|\mathbf{x}_r, i\neq k))\big] \nonumber \\
&- \mathbb{E}_{\hat{\mathbf{z}}\sim p_{noise}, \hat{k}\sim \mathbf{K}}\big[\log (1-p(y_i|G(\hat{\mathbf{z}},\hat{\mathbf{y}}_{\hat{k}}), i<K+1)) \nonumber \\
&+ \log p(y_i|G(\hat{\mathbf{z}},\hat{\mathbf{y}}_{\hat{k}}),i=K+1)\big] \label{eqld}
\end{align}}
\noindent Meanwhile, we maximize $\mathcal{L}_{D}$ to update parameters $\theta_G$ in $G$ in order to compete with $D$. Hence, we can reformulate Eq. \ref{eqld} as a minimization problem $\mathcal{L}_{adv}$:
\begin{align}
\mathcal{L}_{adv} = &- \mathbb{E}_{\hat{\mathbf{z}}\sim p_{noise}, \hat{k}\sim \mathbf{K}}\big[\log p(y_i|G(\hat{\mathbf{z}},\hat{\mathbf{y}}_{\hat{k}}), i=\hat{k}) \nonumber \\
&+ \log (1-p(y_i|G(\hat{\mathbf{z}},\hat{\mathbf{y}}_{\hat{k}}),i\neq\hat{k}))\big] \label{eqla}
\end{align}

In order to to improve the training stability in the A\textlcsc{rt}GAN, we added the L2 pixel-wise reconstruction loss $\mathcal{L}_{L2}$ along with $\mathcal{L}_{adv}$. Given the latent feature $\mathbf{z}$ output from $Enc$ using $\mathbf{x}_r$ as input, $\mathbf{z}$ is fed into $Dec$ to reconstruct the image $\hat{\mathbf{x}}_\mathbf{z}$. Hence, $\mathcal{L}_{L2}$ is defined as:
\begin{equation}
{\mathcal{L}_{L2}} = {\mathbb{E}_{\mathbf{x}_r \sim p_{data}}}
\left[ {\left\| {Dec(Enc(\mathbf{x}_r)) - \mathbf{x}_r} \right\|_2^2} \right]
\end{equation}

\noindent where $\left\|  \cdot  \right\|$ is the second-ordered norm. It should be noted that in the original VAE \cite{larsen2015autoencoding}, $\mathcal{L}_{L2}$ is used to update both the $Enc$ and $Dec$. Conversely, we found that $\mathcal{L}_{L2}$ degrades the quality of the generated images when it is used to update $Enc$. Hence, we only use $\mathcal{L}_{L2}$ when updating $\theta_G$. The final form of the loss function for $G$ is  $\mathcal{L}_{G} = \mathcal{L}_{adv} + \mathcal{L}_{L2}$. Algorithm \ref{pseu} illustrates the training process in our A\textlcsc{rt}GAN model.

\begin{algorithm}[ht]
\caption{Pseudocode for training A\textlcsc{rt}GAN}
\label{pseu}
\begin{algorithmic}[1]
\Require Minibatch size, $n$ and Learning rate, $\lambda$
\Require Randomly initialize ${\theta_D}$ and $\theta_G$ 
\Require Denote parameters of $Dec$, $\theta_{Dec}\in \theta_g$
\While{condition not met}
\State Sample $\hat{\mathbf{Z}}=[\hat{\mathbf{z}}_1,\ldots,\hat{\mathbf{z}}_n]\sim\mathcal{N}(0,1)^{n\times d}$
\State Randomly set $\hat{\mathbf{Y}}_{\hat{k}}=[\hat{\mathbf{y}}_{\hat{k}_1},\ldots,\hat{\mathbf{y}}_{\hat{k}_n}], \hat{k}_i\in\mathbf{K}$
\State Sample minibatch $\mathbf{X}_r=[\mathbf{x}_r^1,\ldots,\mathbf{x}_r^n]$
\State and $\mathbf{k}=[k_1,\ldots,k_n]$
\State $\mathbf{Y} = D(\mathbf{X}_r)$
\State $\hat{\mathbf{X}}=G(\hat{\mathbf{Z}},\hat{\mathbf{Y}}_{\hat{k}})$
\State $\hat{\mathbf{Y}}=D(\hat{\mathbf{X}})$
\State $\theta_D = \theta_D - \lambda\frac{\partial \mathcal{L}_{D}}{\partial \theta_D}$, $\mathcal{L}_{D}\leftarrow \mathbf{Y}, \mathbf{k}, \hat{\mathbf{Y}}, \hat{\mathbf{Y}}_{\hat{k}}$
\State $\mathbf{Z} = Enc(\mathbf{X})$
\State ${\hat{\mathbf{X}}_\mathbf{z}} = Dec(\mathbf{Z})$
\State $\theta_G = \theta_G - \lambda(\frac{\partial \mathcal{L}_{adv}}{\partial \theta_{G}}+\frac{\partial\mathcal{L}_{L2}}{\partial\theta_G})$, $\mathcal{L}_{adv}\leftarrow \hat{\mathbf{Y}}, \hat{\mathbf{Y}}_{\hat{k}}$,\par\hskip\algorithmicindent $\mathcal{L}_{L2}\leftarrow \mathbf{X}, \hat{\mathbf{X}}_z$
\EndWhile
\end{algorithmic}
\end{algorithm}

\section{Experiments}
\label{secE}

\subsection{Dataset}

In this work, we used the publicly available Wikiart dataset\footnote{https://www.wikiart.org/} \cite{saleh2015large} for our experiments. Wikiart is the largest public available dataset that contains around 80,000 annotated artwork in terms of genre, artist and style class. However, not all the artwork are annotated in the 3 respective classes. To be specific, all artwork are annotated for the \textit{style} class. But, there are only 60,000 artwork annotated for the \textit{genre} class, and only around 20,000 artwork are annotated for the \textit{artist} class. We split the dataset into two parts: $30\%$ for testing and the rest for training. 

\subsection{Experiment Settings}
\label{esetting}

In terms of the A\textlcsc{rt}GAN architectures, we used $\alpha=0.2$ for all leaky ReLU. On the other hand, $Dec$ shares the layers Deconv3 to Deconv6 in $G$; and $Enc$ shares the layers Conv1 to Conv4 in $D$. We trained the proposed A\textlcsc{rt}GAN and other models in the experiments for $100$ epochs with minibatch size of 128. For stability, we used the adaptive learning method RmsProp \cite{tieleman2012lecture} for optimization. We set the decay rate to $0.9$ and initial learning rate to $0.001$. We found out that reducing the learning rate during the training process will help in improving the image quality. Hence, the learning rate is reduced by a factor of $10$ at epoch $80$. 

\subsection{Artwork Synthesis Quality}

\begin{figure}[ht]
\centering
\begin{subfigure}{0.1\textwidth}
  \centering
  \includegraphics[width=0.85\textwidth]{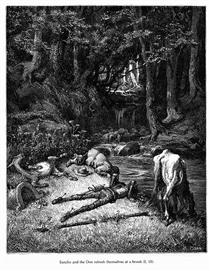}
  \label{Dore1}
\end{subfigure}
\begin{subfigure}{0.36\textwidth}
  \centering
  \includegraphics[height=0.25\textwidth, width=0.98\textwidth]{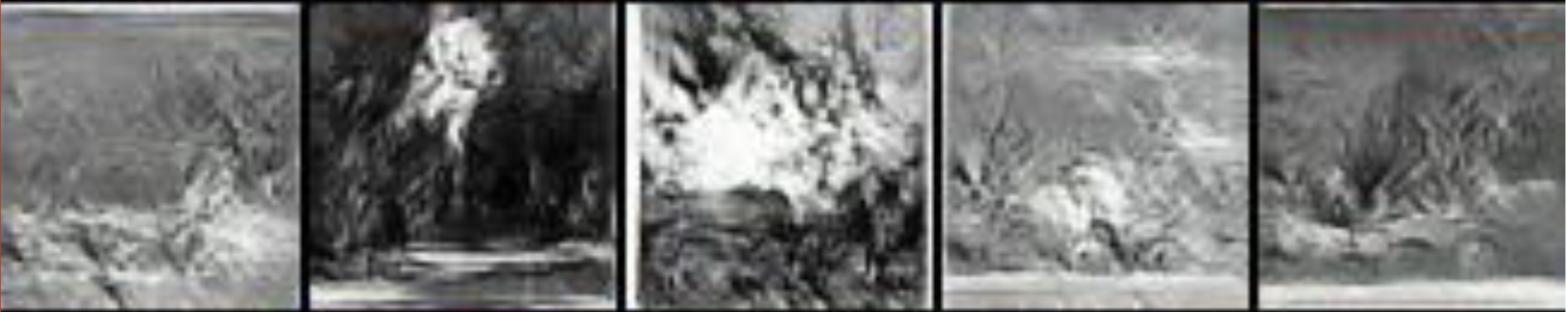}
  \label{Dore2}
\end{subfigure}
\begin{subfigure}{0.1\textwidth}
  \centering
  \includegraphics[height=0.85\textwidth, width=0.95\textwidth]{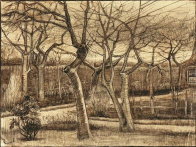}
  \caption{Real}
  \label{vangogh1}
\end{subfigure}
\begin{subfigure}{0.36\textwidth}
  \centering
  \includegraphics[height=0.23\textwidth, width=0.98\textwidth]{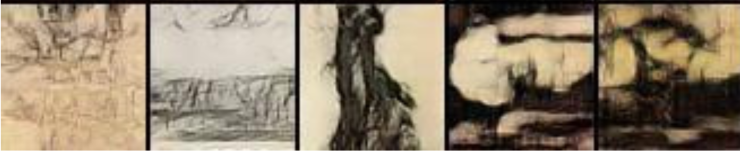}
  \caption{Synthesis}
  \label{vangogh2}
\end{subfigure}
\caption{Sample of the generated \textit{artist} artwork - Gustave Dore (top) and Vincent van Gogh (bottom).}
\label{Gogh}
\end{figure}

\begin{figure}[ht]
\centering
\begin{subfigure}{0.1\textwidth}
  \centering
  \includegraphics[height=1.2\linewidth, width=0.9\textwidth]{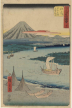}
  \caption{Real}
  \label{unreal11}
\end{subfigure}
\begin{subfigure}{0.36\textwidth}
  \centering
  \includegraphics[height=0.3\textwidth, width=0.9\textwidth]{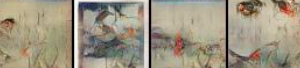}
  \caption{Synthesis}
  \label{unreal21}
\end{subfigure}
\caption{Sample of the generated \textit{style} artwork - Ukiyo-e.}
\label{Ukiyo}
\end{figure}

{\bf Genre:} We compare the quality of the generated artwork trained based on the \textit{genre}. Fig. \ref{fig:clscom} shows sample of the artwork synthetically generated by our proposed A\textlcsc{rt}GAN, DCGAN \cite{radford2015unsupervised} and GAN/VAE, respectively. We can visually notice that the generated artwork from the DCGAN is relatively poor, with a lot of noises (artefacts) in it. In GAN/VAE, we could notice that the generated artwork are less noisy and look slightly more natural. However, we can observe that they are not as compelling. In contrast, the generated artwork from the proposed A\textlcsc{rt}GAN are a lot more natural visually in overall. \\

\noindent {\bf Artist:} Fig. \ref{Gogh} illustrates artwork created by A\textlcsc{rt}GAN based on \textit{artist} and interestingly, the A\textlcsc{rt}GAN is able to recognize the artist's preferences. For instance, most of the {\it Gustave Dore's} masterpieces are completed using engraving, which are usually dull in color as in Fig. \ref{vangogh1}-top. Such pattern was captured and led the A\textlcsc{rt}GAN to draw greyish images as depicted in Figure \ref{vangogh2}-(top). Similarly, most of the Vincent van Gogh's masterpieces in the Wikiart dataset are annotated as {\it Sketch and Study} genre as illustrated in Fig. \ref{vangogh1}-bottom. In this genre, Van Gogh's palette consisted mainly of sombre earth tones, particularly dark brown, and showed no sign of the vivid colours that distinguish his later work, \eg~the famous \textit{\textbf{The Starry Night}} masterpiece. This explains why the artwork synthetically generated by A\textlcsc{rt}GAN is colourless (Fig. \ref{vangogh2}-bottom). \\

\noindent {\bf Style:} Fig. \ref{Ukiyo} presents the artwork synthetically generated by A\textlcsc{rt}GAN based on \textit{style}. One interesting observation can be seen on the \textit{Ukiyo-e} style paintings. Generally, this painting style is produced using the woodblock printing for mass production and a large portion of these paintings appear to be yellowish as shown in Figure \ref{unreal11} due to the paper material. Such characteristic can be seen in the generated \textit{Ukiyo-e} style paintings. Although the subjects in the paintings are hardly recognizable, it is noticeable that A\textlcsc{rt}GAN is trying to mimic the pattern of the subjects.

\subsection{Drawing CIFAR-10 with A\textlcsc{rt}GAN}

\begin{figure}[t]
\centering
\includegraphics[width=0.95\linewidth]{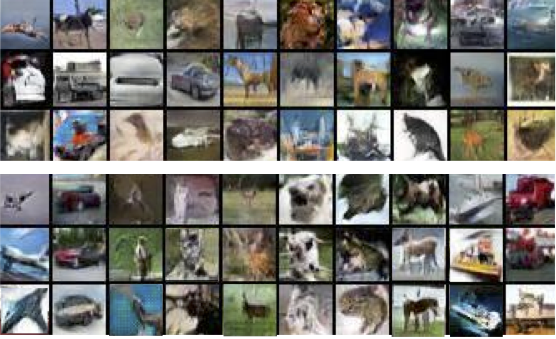}
\caption{Generated CIFAR-10 images using DCGAN \cite{radford2015unsupervised} (top) and A\textlcsc{rt}GAN (bottom).}
\label{fig:cifar}
\end{figure}

\begin{table}[t]
\centering
\caption{Comparison between different GAN models using log-likehood measured by Parzen-window estimate.}
\label{parzen}
\begin{tabular}{cc}
Model & Log-likelihood \\
\hline
DCGAN \cite{radford2015unsupervised} & $2348\pm67$ \\
GAE/VAE & $2483\pm67$ \\
\textbf{A\textlcsc{rt}GAN} & $\boldsymbol{2564\pm67}$
\end{tabular}
\end{table}

\begin{figure}[t]
\centering
\includegraphics[height=0.3\linewidth, width=0.95\linewidth]{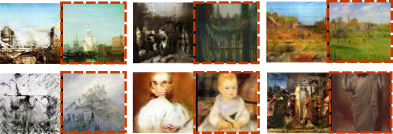}
\caption{Nearest neighbour comparisons. Paintings in the red dotted boxes are the corresponding nearest paintings.}
\label{fig:nearest}
\end{figure}

We trained both the DCGAN \cite{radford2015unsupervised} and A\textlcsc{rt}GAN to generate natural images using the CIFAR-10 dataset. The generated samples on CIFAR-10 are presented in Fig. \ref{fig:cifar}. As aforementioned, the DCGAN is able to generate much recognizable images, contrast to its failure in generating artwork. This implies our earlier statements that the objects in CIFAR-10 have a fairly structured shape, and so it is much easier to learn compared to the artwork that are abstract. Even so, we could still notice some of the generated shapes are not as compelling due to CIFAR-10 exhibits huge variability in shapes compared to CUB-200 dataset of birds and LFW dataset of face. Meanwhile, we can observed that the proposed A\textlcsc{rt}GAN is able to generate much better images. For instance, we can see the auto-mobile and horse with clear shape.

\subsection{Quantitative Analysis}

By using the GAN models trained previously, we measure the log-likelihood of the generated artwork. Following Goodfellow \etal~\cite{goodfellow2014generative}, we measure the log-likehood using the Parzen-window estimate. The results are reported in Table \ref{parzen} and show that the proposed A\textlcsc{rt}GAN performs the best among the compared models. However, we should note that these measurements might be misleading \cite{theis2015note}. In addition, we also find the nearest training examples of the generated artwork by using exhaustive search on L2 norm in the pixel space. The comparisons are visualized in Fig. \ref{fig:nearest} and it shows that the proposed A\textlcsc{rt}GAN does not simply memorize the training set. 

\section{Conclusions}
\label{secC}

In this work, we proposed a novel A\textlcsc{rt}GAN to synthesize much challenging and complex images. In the empirical experiments, we showed that the feedback from the label information during the back-propagation step improves the quality of the generated artwork. A natural extension to this work is to use a deeper A\textlcsc{rt}GAN to encode more detail concepts. Furthermore, we are also interested in jointly learn these modes, so that A\textlcsc{rt}GAN can create artwork based on the combination of several modes.

\newpage
{\small
\bibliographystyle{ieee}
\bibliography{egbib}
}

\section{Appendix}


\subsection{Model Configurations}

Table \ref{generator} and \ref{discriminator} list the detailed configurations of the Generator $G$ and Discriminator $D$ in our proposed A\textlcsc{rt}GAN model. 

\begin{table}[ht]
\begin{minipage}{0.9\linewidth}
\centering
\caption{Generator}
\resizebox{\columnwidth}{!}{%
\begin{tabular}{c|c|c|c|c|c|c}
layer & \# of filters & filter size & strides & paddings & batchnorm & activation \\
\hline
Deconv1 & 1024 & 4 & 1 & 0 & yes & ReLU \\
Deconv2 & 512 & 4 & 2 & 1 & yes & ReLU \\
Deconv3 & 256 & 4 & 2 & 1 & yes & ReLU \\
Deconv4 & 128 & 4 & 2 & 1 & yes & ReLu \\
Deconv5 & 128 & 3 & 1 & 1 & yes & ReLu \\
Deconv6 & 3 & 4 & 2 & 1 & no & Sigmoid
\label{generator}
\end{tabular}%
}
\end{minipage}%
\quad
\begin{minipage}{0.9\linewidth}
\centering
\caption{Discriminator}
\resizebox{\columnwidth}{!}{%
\begin{tabular}{c|c|c|c|c|c|c}
layer & \# of filters & filter size & strides & paddings & batchnorm & activation \\
\hline
Conv1 & 128 & 4 & 2 & 1 & no & leakyReLU \\
Conv2 & 128 & 3 & 1 & 1 & yes & leakyReLU \\
Conv3 & 256 & 4 & 2 & 1 & yes & leakyReLU \\
Conv4 & 512 & 4 & 2 & 1 & yes & leakyReLU \\
Conv5 & 1024 & 4 & 2 & 1 & yes & leakyReLU \\
fc6 & 1 & - & - & - & no & Sigmoid
\label{discriminator}
\end{tabular}%
}
\end{minipage}%
\end{table}

\subsection{More Results for Wikiart Dataset}

\begin{figure*}[ht]
\centering
\includegraphics[width=\textwidth]{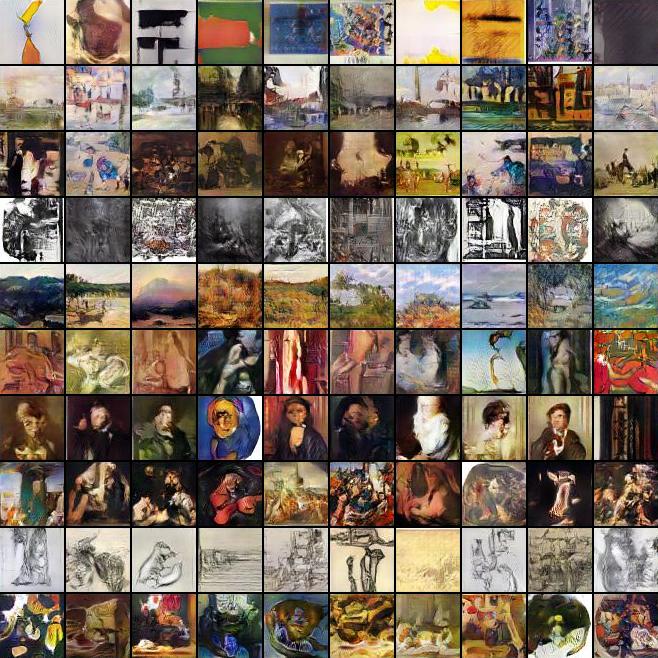}
  \caption{{\bf A\textlcsc{rt}GAN:} Generated artwork based on the {\it genre} class. From top to bottom: (1) Abstract Painting, (2) Cityscape, (3) Genre Painting, (4) Illustration, (5) Landscape, (6) Nude Painting, (7) Portrait, (8) Religious Painting, (9) Sketch and Study, and (10) Still Life.}
  \label{genccgan}
\end{figure*}

\begin{figure*}[htp]
\centering
\includegraphics[width=0.97\linewidth]{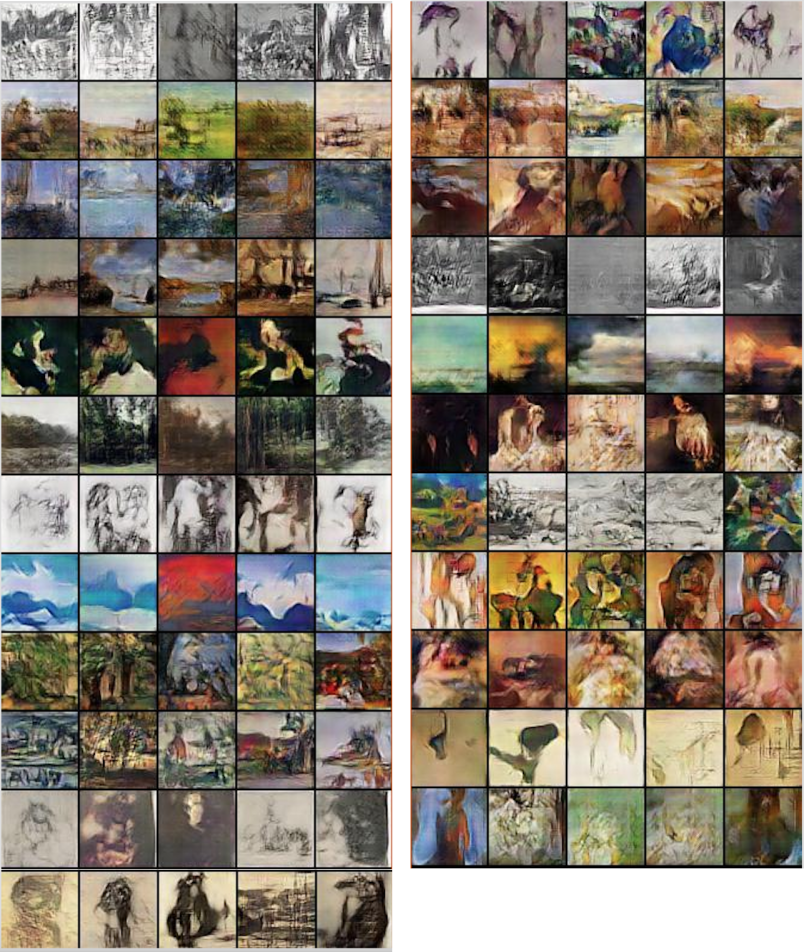}
\caption{{\bf A\textlcsc{rt}GAN:} Generated artwork based on {\it artists} class. ({\bf Left}) From top to bottom: (1) Albrecht Durer, (2) Camille Pissarro, (3) Claude Monet, (4) Eugene Boudin, (5) Ilya Repin, (6) Ivan Shishkin, (7) Marc Chagall, (8) Nicholas Roerich, (9) Paul Cezanne, (10) Pyotr Konchalovsky, (11) Rembrandt, (12) Vincent van Gogh. ({\bf Right}) From top to bottom: (1) Boris Kustodiev, (2) Childe Hassam, (3) Edgar Degas, (4) Gustave Dore, (5) Ivan Aivazovsky, (6) John Singer Sargent, (7) Martiros Saryan, (8) Pablo Picasso, (9) Pierre Auguste Renoir, (10) Raphael Kirchner, (11) Salvador Dali.}
\label{fig:artist}
\end{figure*}

In Figure \ref{genccgan}-\ref{fig:artist}, we show more results on the {\it genre} and {\it artist} class. For instance, {\it Nicholas Roerich} had travelled to many Asia countries and finally settled in the Indian Kullu Valley in the Himalayan foothills. Hence, he has many paintings that are related to mountain using \textbf{Symbolism} style\footnote{instead of emphasizing on realistic, \textbf{Symbolism} depicts the subjects using forms, lines, shapes, and colors}. This can be seen in the generated paintings (Figure \ref{fig:artist}, no. 8 from left) which look like mountain even-though unrealistic. On another example, A\textlcsc{rt}GAN also shows that {\it Ivan Shishkin}'s persistent in drawing forest landscape paintings (Figure \ref{fig:artist}, no. 6 from left). {\it Ivan Shishkin} is one of the most prominent Russian landscape painters. By his contemporaries, Shishkin was given the nicknames ``Titan of the Russian Forest", ``Forest Tsar", ``Old Pine Tree" and ``Lonely Oak" as there was no one at that time who depicted trees more realistically, honestly and with greater love.

\begin{figure*}[htp]
\centering
\begin{subfigure}{0.48\textwidth}
  \centering
  \includegraphics[width=\textwidth]{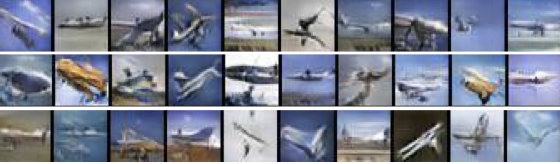}
  \caption{Airplane}
  \label{airplane}
\end{subfigure}
\quad
\begin{subfigure}{0.48\textwidth}
  \centering
  \includegraphics[width=\textwidth]{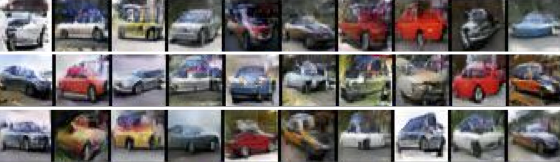}
  \caption{Automobile}
  \label{automobile}
\end{subfigure}
\begin{subfigure}{0.48\textwidth}
  \centering
  \includegraphics[width=\textwidth]{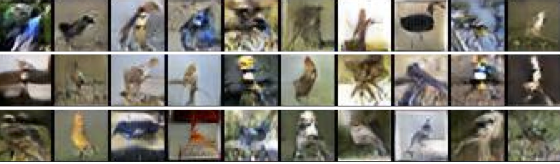}
  \caption{Bird}
  \label{bird}
\end{subfigure}
\quad
\begin{subfigure}{0.48\textwidth}
  \centering
  \includegraphics[width=\textwidth]{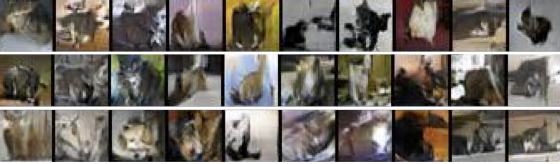}
  \caption{Cat}
  \label{cat}
\end{subfigure}
\begin{subfigure}{0.48\textwidth}
  \centering
  \includegraphics[width=\textwidth]{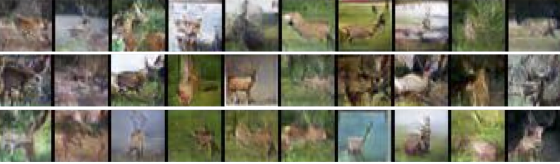}
  \caption{Deer}
  \label{deer}
\end{subfigure}
\quad
\begin{subfigure}{0.48\textwidth}
  \centering
  \includegraphics[width=\textwidth]{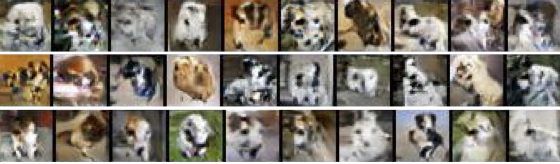}
  \caption{Dog}
  \label{dog}
\end{subfigure}
\begin{subfigure}{0.48\textwidth}
  \centering
  \includegraphics[width=\textwidth]{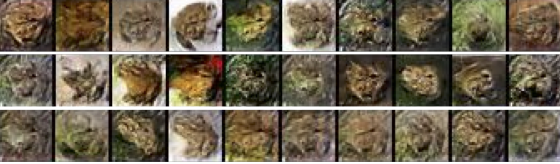}
  \caption{Frog}
  \label{frog}
\end{subfigure}
\quad
\begin{subfigure}{0.48\textwidth}
  \centering
  \includegraphics[width=\textwidth]{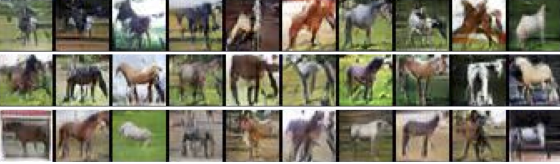}
  \caption{Horse}
  \label{horse}
\end{subfigure}
\begin{subfigure}{0.48\textwidth}
  \centering
  \includegraphics[width=\textwidth]{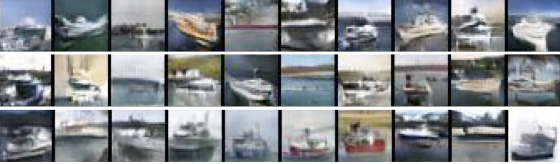}
  \caption{Ship}
  \label{ship}
\end{subfigure}
\quad
\begin{subfigure}{0.48\textwidth}
  \centering
  \includegraphics[width=\textwidth]{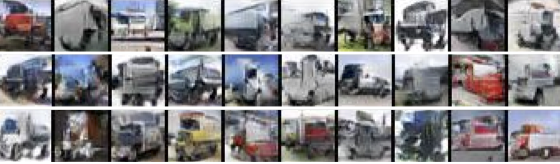}
  \caption{Truck}
  \label{truck}
\end{subfigure}
\caption{Samples of the generated CIFAR-10 using the proposed A\textlcsc{rt}GAN, and we can see clear shape in every generated images of respective classes.}
\label{cifar10}
\end{figure*}

\subsection{More Results on CIFAR-10}

In this section, we report more results on the CIFAR-10 dataset. Figure \ref{cifar10} shows the generated images in each of the class. Even though the objects in CIFAR-10 exhibit huge variability in shapes, we can see that A\textlcsc{rt}GAN is still able to generate object-specific appearances and shapes.

\subsection{More Results on Neighbourhood}

\begin{figure*}[htp]
\centering
\begin{subfigure}{0.30\textwidth}
  \centering
  \includegraphics[width=\textwidth]{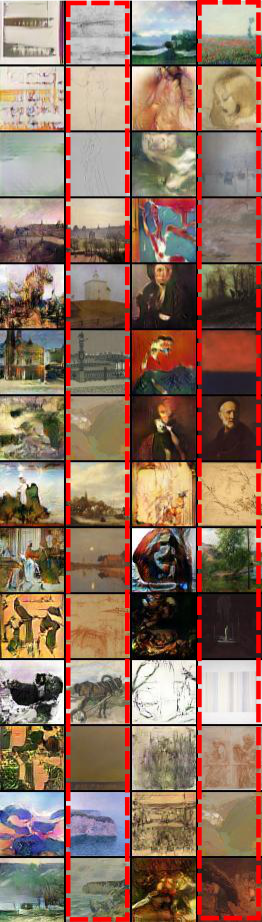}
  \caption{Genres}
  \label{neargenre}
\end{subfigure}
\quad
\begin{subfigure}{0.30\textwidth}
  \centering
  \includegraphics[width=\textwidth]{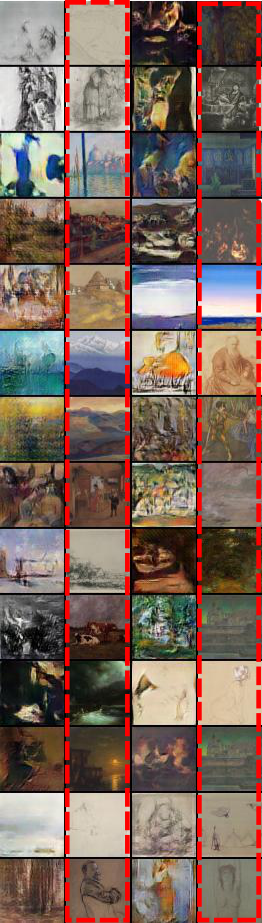}
  \caption{Artist}
  \label{nearartist}
\end{subfigure}
\quad
\begin{subfigure}{0.30\textwidth}
  \centering
  \includegraphics[width=\textwidth]{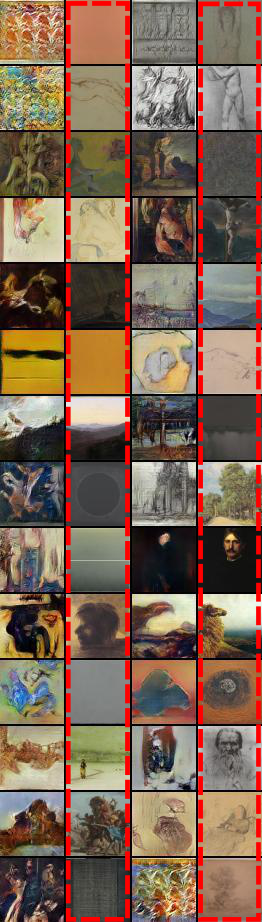}
  \caption{Style}
  \label{nearstyle}
\end{subfigure}
\caption{Samples of the generated paintings and their nearest neighbours (red dotted box) in the Wikiart dataset. It can be noticed that the proposed A\textlcsc{rt}GAN does not simply memorize the training set.}
\label{near}
\end{figure*}

In Figure \ref{near}, we show more examples of the nearest neighbour of the generated paintings. These examples justify that the proposed A\textlcsc{rt}GAN does not simply memorize the training set. 

\end{document}